\documentclass[conference]{IEEEtran}
\IEEEoverridecommandlockouts

\usepackage{cite}
\usepackage{amsmath,amssymb,amsfonts}
\usepackage{algorithmic}
\usepackage{graphicx}
\usepackage{textcomp}
\usepackage{xcolor}

\usepackage{subcaption}
\usepackage{tikz}
\usetikzlibrary{shadows, arrows.meta, positioning}
\usepackage{fontawesome5} 
\usepackage{graphicx,amssymb,amsmath,mathrsfs,bbm}
\usepackage{booktabs}
\usepackage{tcolorbox}
\usepackage{orcidlink}

\newcommand{\maj}{\mathrm{maj}}

\usepackage{graphicx}
\usepackage{tabularx} 
\def\BibTeX{{\rm B\kern-.05em{\sc i\kern-.025em b}\kern-.08em
    T\kern-.1667em\lower.7ex\hbox{E}\kern-.125emX}}
\begin{document}

\title{Training Dynamics of Neural Software Defect Predictors under Coupled Data-Quality Issues\\
}

\author{\IEEEauthorblockN{1\textsuperscript{st} Emmanuel C. Dapaah}
\IEEEauthorblockA{\textit{Institute of Computer Science} \\
\textit{University of Goettingen}\\
Goettingen, Germany \\
dapaah@cs.uni-goettingen.de}
\and
\IEEEauthorblockN{2\textsuperscript{nd} Philip Makedonski}
\IEEEauthorblockA{\textit{Institute of Computer Science} \\
\textit{University of Goettingen}\\
Goettingen, Germany \\
makedonski@informatik.uni-goettingen.de}
\and
\IEEEauthorblockN{3\textsuperscript{rd} Jens Grabowski}
\IEEEauthorblockA{\textit{Institute of Computer Science} \\
\textit{University of Goettingen}\\
Goettingen, Germany \\
grabowski@informatik.uni-goettingen.de}
}

\maketitle

\begin{abstract}
\textbf{Context---} 
Software defect prediction supports maintenance decisions such as testing prioritization, release-risk assessment, and quality monitoring. However, metric-based SDP datasets often contain coupled data-quality issues, especially class imbalance and class overlap. Prior work has mainly measured their impact through endpoint performance, while recent evidence suggests that such issues may also appear in neural training dynamics (gradients, weights, biases, error trajectories). However, these studies examine issues in isolation, leaving open how internal neural network training patterns manifest when data quality issues are coupled.

\textbf{Objective---} We investigate how training-dynamics patterns from class imbalance, overlap, and their coupling can be characterized under interaction-aware conditions in deep learning-based SDP.

\textbf{Method---} We conduct a controlled intervention study on class-level UBD datasets\cite{ferenc_public_2018}, training a fixed MLP under imbalance-only, overlap-only, and joint conditions across five seeds. Training dynamics are logged per epoch; fidelity is monitored via coupling ratios. Patterns are characterized using effect sizes, trajectories, sensitivity analyses, and rule-based classification.

\textbf{Expected contribution---} The study will produce an interaction-aware empirical protocol and a candidate taxonomy of training-dynamics patterns for coupled data-quality issues in metric-based SDP.
\end{abstract}


\section{Introduction}
Software defect prediction (SDP) estimates which software artifacts are likely to contain defects\cite{li_software_2024} and supports testing prioritization, release-risk assessment, inspection allocation, and defect management. When maintenance teams use defect-prediction outputs to allocate limited quality-assurance effort, the question is not only whether a model achieves high predictive performance, but whether its behaviour can be trusted under the data conditions commonly found in software repositories.

A persistent threat is training-data quality. Metric-based SDP datasets often exhibit class imbalance (defective instances are rarer)\cite{liu_comparative_2022,k_j_impact_2022} and class overlap (defective and non-defective instances share similar feature-space regions)~\cite{gong_comprehensive_2023,zhang2026}. These issues can reduce performance, especially for the minority class, but their practical consequences extend beyond performance degradation. If a defect-prediction model behaves poorly, developers and researchers need to know whether the problem is likely caused by skewed class representation, ambiguous class boundaries, or their interaction. Without this information, data cleaning, model debugging, and maintenance-decision support remain largely trial-and-error.

Most SDP studies evaluate data-quality issues using endpoint metrics such as AUC, F1-score, recall, precision, or balanced accuracy~\cite{Kabir2019,tantithamthavorn2020,liu_comparative_2022,bhandari_data_2023,eberlein_effect_2025}. These metrics show whether performance degrades, but not how the model learned or which data issue shaped learning. For example, two datasets may produce similar F1-score degradation even though one model struggled because defective instances were underrepresented, while another struggled because defective and non-defective instances were difficult to separate. In such cases, endpoint performance alone gives little guidance on whether practitioners should rebalance the data, reduce boundary ambiguity, revise labels, or treat the model as unreliable for the intended maintenance decision.

Recent work suggests data-quality problems may be reflected in training dynamics (gradients, weights, biases, errors)~\cite{deepfd_2022,shah2025}, offering earlier diagnostic signals. For SDP, this opens a direction where training behaviour may reveal sensitivity to particular data conditions. However, this research direction is difficult because class imbalance and class overlap are not independent in practice. Changing class proportions through sampling may also change neighbourhood structure and apparent boundary ambiguity. Conversely, moving or removing boundary-near instances to modify overlap may also affect class proportions. Therefore, a training-dynamics pattern observed under an intervention planned to manipulate imbalance or overlap may not reflect the intended issue alone. It may also reflect unintended changes in the other issue or an interaction between the two. This creates a methodological limitation in existing studies using training dynamics as evidence for data-quality diagnosis in neural SDP models.

This study addresses that limitation by proposing an interaction-aware intervention study of class imbalance, class overlap, and their coupling in metric-based SDP. We will use class-level SDP datasets from the Unified Bug Dataset\cite{ferenc_public_2018}, apply controlled imbalance-only, overlap-only, joint imbalance--overlap, and progressive-reduction conditions, and train a fixed multilayer perceptron under repeated random seeds. After every intervention, we will recompute realized class imbalance and overlap metric values to assess whether the intended data-quality issue changed and whether the non-target issue drifted. We will then analyze predefined training-dynamics metric families, including error dynamics, gradient magnitude and propagation, gradient distribution shape, and weight and bias summaries.

The study aims to catalog candidate training-dynamics patterns associated with imbalance, overlap, and their coupling, classify them as issue-associated, shared, or joint-condition patterns, and provide a reproducible SDP training-dynamics protocol. This is a necessary step toward practical data-quality debugging for neural defect-prediction pipelines.

\section{Related Work}

\subsection{Software Defect Prediction and Data Quality}
Public SDP datasets such as PROMISE and UBD have enabled comparative defect-prediction research~\cite{caglayan2012,ferenc_public_2018}. Prior work shows that model conclusions are sensitive to dataset properties, learner choice, evaluation metrics, and preprocessing~\cite{hall_systematic_2012,shepperd2014,tantithamthavorn2020}, making data-quality issues such as imbalance and overlap important for both performance and interpretation.

\subsection{Class Imbalance, Class Overlap, and Their Coupling}
Class imbalance is widely studied in SDP because defective instances are often rare, motivating SMOTE and undersampling~\cite{chawla_smote_2002,yen2006}. Class overlap has also gained attention because defective and non-defective modules may share similar metric values~\cite{gong_comprehensive_2023, dapaah2025dataquality}. More recent work suggests that imbalance and overlap should not always be treated as independent data-quality issues as rebalancing can alter neighbourhood structure, overlap handling can affect class proportions, and combined treatments may change both performance and interpretation~\cite{wang2024,zhang2026}. These studies motivate our focus on coupling. However, most existing work still evaluates imbalance and overlap primarily through endpoint performance measures~\cite{prati_2004,tantithamthavorn2020,dapaah2025dataquality} or feature-importance stability. In contrast, our study will examine whether training-dynamics patterns vary when imbalance and overlap are manipulated separately, jointly, and under monitored non-target drift.

\subsection{From Endpoint Metrics to Training Dynamics}
Endpoint metrics are necessary for evaluating defect predictors~\cite{tantithamthavorn2020,bhandari_data_2023}, but they do not show how neural models learn under problematic data conditions. Recent data-bug work in software engineering uses training logs, gradients, weights, biases, and trajectories to study data-quality effects in deep learning~\cite{shah2025}. This line of work provides an important motivation for treating training dynamics as possible evidence of data-condition sensitivity. Nevertheless, we are not aware of prior SDP work that combines single-issue, joint, and drift-monitored intervention protocols for candidate training-dynamics classification. Our study addresses this gap by measuring the realized severity of imbalance and overlap after each intervention, monitoring non-target drift, and applying progressive reduction and rule-based classification to interpret candidate patterns.

\section{Research Questions}

Our investigation is guided by three open research questions:

\textbf{RQ1. How do candidate training-dynamics patterns change when class imbalance and class overlap are manipulated separately under controlled, fidelity-monitored interventions?}

\emph{Rationale.} This RQ will characterize severity-response patterns under imbalance-only and overlap-only intervention paths. We will not assume that either issue is perfectly isolated; instead, we will measure the realized severity of both imbalance and overlap after every intervention and interpret patterns in light of any non-target drift.

\textbf{RQ2. How do candidate training-dynamics patterns vary when imbalance and overlap are introduced jointly?}

\emph{Rationale.} This RQ examines whether single-issue patterns appear together, weaken, dominate, or change under joint intervention. Although C2 and N1 are both bounded in [0,1], they measure different constructs. Therefore, we will use low, medium, and high levels to structure the joint intervention, but we will interpret the results using the realized post-intervention imbalance and overlap measurements rather than assuming direct equivalence between the two metrics.

\textbf{RQ3. How do candidate patterns change under progressive issue reduction?}

\emph{Rationale.} Single-direction intervention alone may produce patterns that are sensitive to the intervention mechanism. This RQ uses progressive reduction as triangulation: patterns are more credible if they change under injection and weaken under reduction. We do not interpret this as causal reversal, but as evidence about pattern robustness.

\section{Datasets and Preprocessing}

\subsection{Dataset Source and Unit of Analysis}

We will use the Unified Bug Dataset (UBD), a public curated benchmark for Java bug prediction that consolidates metric-based datasets from five public sources and recomputes a common set of source-code metrics~\cite{ferenc_public_2018}. UBD provides both class-level and file-level versions. To avoid mixing granularities and to reduce duplicate representations of the same systems, the primary analysis will use the \emph{class-level} UBD datasets only. Each dataset will be treated as an independent binary SDP task in which the unit of prediction is a Java class and the target indicates whether the class is defective. File-level UBD datasets will not be mixed into the primary analysis; if used at all, they will be reported separately as a robustness extension rather than as part of the main dataset pool.

We will start from the full set of candidate class-level UBD CSV files available in our dataset pool. If duplicate representations of the same project, version, and granularity are present, we will retain one canonical UBD representation and remove exact duplicates. Different releases or versions of the same project will be retained as separate datasets because they represent different prediction contexts. 

\subsection{Inclusion and Exclusion Criteria}

The inclusion and exclusion criteria will be applied before model training and before inspecting any training-dynamics outcomes. A dataset will be included in the primary analysis only if it is a class-level UBD CSV dataset with a binary target variable, numeric or Boolean predictors, at least 200 instances before splitting, and at least 50 original instances in each class. Identifier columns and non-feature metadata will be removed. Labels must be encodable as 0 and 1, and stratified train/validation/test splitting must preserve both classes in all partitions.

Each intervention condition must retain at least 20 training instances per class after injection or progressive reduction and must contain only finite feature values. Entirely missing, constant, or non-finite feature columns will be removed; if this makes a dataset or condition invalid, the affected dataset or condition will be excluded and logged with an explicit reason. If a dataset is valid for some intervention families but invalid for others, it will be retained only for the RQ-specific analyses for which the required intervention conditions are valid. Based on the current UBD candidate pool and script assumptions, we expect approximately 40--50 datasets to remain eligible, but the final number will be determined only after applying these registered filtering rules.

The 200-instance and 50-per-class pre-intervention floors make the 20-per-class post-intervention floor achievable after severe undersampling or editing; the latter is the minimum we consider viable for mini-batch training and nearest-neighbour overlap measurement. Sensitivity will vary this floor to 15 and 25.

\subsection{Splitting and Preprocessing}

Each dataset will be processed independently using a fixed stratified split with seed 42: 20\% test, then 20\% of the remaining data for validation, yielding approximately 64/16/20 train/validation/test partitions. The same validation and test sets will be reused across interventions and seeds. Median imputation and \texttt{StandardScaler} will be fitted on the original training partition only and applied to all partitions; Boolean features will be encoded as 0/1 before scaling. Interventions will modify only transformed training data, never validation or test data. This design ensures that differences across intervention conditions reflect changes in the training data rather than changes in the evaluation data or data-leakage from validation/test partitions.

\section{Execution Plan}

\subsection{Data-Quality Measurement}

For each dataset and intervention condition, we will compute two data-quality measures on the preprocessed training data only. Class imbalance will be measured using C2, the imbalance-ratio complexity measure~\cite{lorena2020}. For $n_c$ classes, $n_i$ instances in class $i$, and $n$ total instances, we compute:

\[
IR = \frac{n_c - 1}{n_c}\sum_{i=1}^{n_c}\frac{n_i}{n-n_i},
\qquad
C2 = 1 - \frac{1}{IR}.
\]

Class overlap will be measured using N1, the fraction of borderline points~\cite{lorena2020}. We will compute N1 using Euclidean distances over standardized training features. Let $G=(V,E)$ be the minimum spanning tree over the training instances, where each vertex $i \in V$ represents one training instance $x_i$ with class label $y_i$, and each edge $(i,j)\in E$ connects two instances in the tree. N1 is the fraction of instances that are incident to at least one MST edge connecting opposite-class instances:

\[
N1 =
\frac{
\left|\left\{i \in V \mid \exists j \in V : (i,j)\in E \land y_i \neq y_j \right\}\right|
}{n}.
\]

Here, $n=|V|$ is the number of training instances. C2 will be the target metric for imbalance interventions, and N1 will be the target metric for overlap interventions. Both metrics will be recomputed after every intervention condition. The non-target metric will be used to monitor drift; for example, N1 drift during imbalance intervention and C2 drift during overlap intervention. Although both metrics are bounded between 0 and 1, they measure different constructs, so equal numeric values will not be treated as equivalent severity.

\subsection{Model Architecture and Training Protocol}

We will use a fixed feed-forward multilayer perceptron (MLP) for all datasets and intervention conditions. The input size will match the number of preprocessed predictors; hidden layers will have 128, 64, and 32 units with ReLU activations and dropout ($p=0.2$); the output layer will produce two logits. Weights will be initialized with Kaiming normal initialization and biases with zeros. Models will be trained in PyTorch using cross-entropy loss, Adam ($lr=10^{-3}$, weight decay $10^{-4}$), shuffled mini-batches of size 64, and 100 fixed epochs. We will not use class weights, class-balanced batch sampling, or early stopping in the primary analysis, because these mechanisms could partially counteract the intended imbalance interventions or make trajectories incomparable across conditions. Dropout will be active during training and disabled during validation/test evaluation. We keep the architecture fixed to avoid confounding data-quality intervention effects with model-selection decisions.

Each intervention-generated training set will be trained with seeds $\{42,43,44,45,46\}$. Training and validation metrics will be logged each epoch; test metrics will be computed once after training. Test metrics are reported only as endpoint context and are not used for candidate pattern classification. Gradient summaries will be recorded after backpropagation and before the optimizer update, then aggregated to epoch-level summaries. We will fix Python, NumPy, and PyTorch seeds, enable deterministic PyTorch settings where feasible, and report package versions, CUDA/device information, and hardware in the artifact package.

As a robustness check, we train a secondary [256-128-64] MLP under identical optimization, regularization, and hyperparameter settings, isolating architectural depth as the variable of interest. Future work may examine interactions between training configuration and data-quality symptoms.

\subsection{Fault-Injection Protocols}

\subsubsection{Class Imbalance Fault-Injection Procedure}

The imbalance-injection protocol will construct a controlled low-imbalance reference and then progressively reduce the minority class. The reference will be created by random oversampling with replacement on the preprocessed training data only, using random state 123, until the minority class matches the majority-class count. This reference will not be treated as natural or clean; it is only a controlled starting point for manipulating class ratios.

Starting from this reference, we will keep the majority class fixed and undersample without replacement from the oversampled minority pool. For each severity level
\[
\alpha \in \{0.00,0.25,0.50,0.75,1.00\},
\]
the target majority/minority ratio will be
\[
r(\alpha)=1+\alpha(5-1),
\]
where $\alpha=0$ corresponds to the low-imbalance reference and $\alpha=1$ corresponds to an intended 5:1 majority/minority ratio. The 5:1 upper bound is chosen to create a clear imbalance gradient while avoiding extreme minority depletion that would make stratified evaluation, nearest-neighbour measures, and neural training unstable.

For each severity level, the target minority count will be
\[
n_{\min}(\alpha)=\max\left(\left\lceil \frac{n_{\maj}}{r(\alpha)} \right\rceil, 20\right),
\]
where $n_{\maj}$ is the fixed majority-class count. The floor of 20 training instances per class is used to avoid intervention conditions that are too small for stratified evaluation, neighbourhood-based overlap measurement, and neural-network training. Conditions that cannot satisfy this floor will be marked invalid for the corresponding analysis. After every imbalance condition, we will recompute realized imbalance (C2) and overlap (N1) measurements to assess intervention fidelity and non-target drift.

\subsubsection{Class Overlap Fault-Injection Procedure}

The overlap-injection protocol will be treated as a controlled boundary-ambiguity stressor, not as a claim of naturally occurring overlap. We will first construct a lower-overlap reference using Repeated Edited Nearest Neighbours (RENN) on standardized training~\cite{enn1976}. The method parameters will be fixed before execution and will not be tuned per dataset. If RENN removes too many instances, leaves a single-class training set, or violates the minimum class floor, the dataset will be marked invalid for overlap-injection analysis.

Starting from the lower-overlap reference, we will identify boundary-near candidates separately within each class. For each instance $x_i$, we will find its nearest opposite-class neighbour $x_{opp(i)}$ using Euclidean distance over standardized training features. Within each class, the closest 25\% of instances to an opposite-class neighbour will form the candidate pool. This class-wise rule prevents the candidate pool from being dominated by the majority class while focusing the perturbation on boundary-near regions.

For each severity level
\[
\alpha \in \{0.00,0.25,0.50,0.75,1.00\},
\]
we will modify a nested fraction $\alpha$ of the candidate pool. A selected instance will be moved toward its nearest opposite-class neighbour while retaining its original label:
\[
x_i'=(1-\gamma)x_i+\gamma x_{opp(i)}, \quad \gamma=0.80.
\]
The value $\gamma=0.80$ is used as a strong but bounded perturbation: it moves selected samples close to the opposite-class region without replacing them by the opposite-class instance. Because labels are preserved, the procedure may introduce label--feature tension; we therefore interpret it as a boundary-ambiguity stressor and discuss this as a construct-validity threat.

After every overlap condition, we will recompute realized overlap (N1) and imbalance (C2) measurements. A dataset will be included in the primary overlap-injection analysis only if realized overlap increases with severity according to the intervention-fidelity check; otherwise, it will be flagged for sensitivity analysis or excluded from the overlap-specific analysis. Conditions that produce non-finite values or violate the minimum class floor will be marked invalid. We will also run a sensitivity analysis with $\gamma=0.60$ to assess whether candidate patterns depend on the chosen perturbation strength.

\subsubsection{Joint Imbalance--Overlap Intervention}

The joint intervention addresses RQ2 by examining candidate training-dynamics patterns when imbalance and overlap are introduced together. We will use a factorial intervention design\cite{Mukerjee2006}, crossing three nominal imbalance levels with three nominal overlap levels:
\[
\alpha_I,\alpha_O \in \{0.00,0.50,1.00\}.
\]
This yields nine joint conditions per dataset. The levels are nominal design levels only; they do not imply equal substantive severity across imbalance and overlap. Therefore, after each joint condition, we will recompute the realized imbalance and overlap measurements and use these realized values, rather than nominal $\alpha$ labels alone, in the analysis.

The primary intervention order will be imbalance first, followed by overlap, because the overlap perturbation preserves labels and is therefore less likely to directly change the class-count ratio after imbalance has been established. For each joint condition, we will record nominal levels, realized data-quality measurements, class counts, modified samples, and non-target drift. Conditions violating the minimum class floor, producing non-finite values, or failing fidelity checks will be marked invalid. As an order-sensitivity check, we will repeat the joint intervention in reverse order for up to ten retained datasets selected to cover low, medium, and high original imbalance/overlap profiles where available.

\subsection{Progressive Reduction Interventions}

Progressive reduction addresses RQ3 by providing convergent exploratory evidence about whether candidate patterns change when the targeted data-quality issue is reduced. We will not describe these procedures as cleaning, reversal, or causal validation, because both procedures introduce their own artifacts. Instead, they are treated as additional interventions that reduce the measured severity of imbalance or overlap under a fixed protocol.

For imbalance reduction, we will start from the original preprocessed training partition and progressively increase the minority class toward the majority-class count using SMOTE~\cite{chawla_smote_2002}. For each reduction level
\[
\alpha \in \{0.00,0.25,0.50,0.75,1.00\},
\]
the target minority count will be
\[
n_{\mathrm{min}}'(\alpha)
=
n_{\mathrm{min}}
+
\left\lfloor
\alpha (n_{\mathrm{maj}}-n_{\mathrm{min}})
\right\rfloor ,
\]
where $n_{\mathrm{min}}$ and $n_{\mathrm{maj}}$ are the original minority- and majority-class counts in the training partition. SMOTE will use $k=5$ neighbours and random state 123; if the minority class is too small, $k$ will be reduced to $n_{\mathrm{min}}-1$. The majority class will remain unchanged. Each condition must retain at least 20 training instances per class.

For overlap reduction, we will use a RENN-based progressive editing procedure on standardized training features~\cite{enn1976}. RENN will first identify candidate majority-class instances located in locally ambiguous neighbourhoods. These candidates will be ranked by boundary ambiguity, operationalized by their distance to the nearest opposite-class neighbour and local class disagreement. For each $\alpha$, we will remove a nested fraction $\alpha$ of these candidates, while preserving the minimum floor of 20 training instances per class. Conditions that remove too many samples, produce a single-class training set, or generate non-finite values will be marked invalid.

After every progressive reduction condition, we will recompute realized imbalance (C2) and overlap (N1) measurements. Progressive reduction is used as a triangulation step, not as causal reversal. A candidate pattern is considered more credible when its direction under injection is directionally opposed under the corresponding reduction intervention, provided non-target drift remains below the pre-specified threshold.

\subsection{Intervention-Fidelity and Non-Target Drift Checks}
After each intervention condition, we will recompute the realized imbalance and overlap metrics on the modified training data. A single-issue intervention path will be treated as fidelity-consistent only if the target metric changes in the expected direction with Spearman's $|\rho|\geq0.70$ and changes by at least 0.10 between its lowest and highest realized values across severity levels. Non-target drift will be flagged when the non-target metric changes by at least 0.10 relative to the reference condition. We will also compute a coupling ratio,
\[
CR=\frac{|\Delta N|}{|\Delta T|+10^{-6}},
\]
where $\Delta N$ and $\Delta T$ are the non-target and target metric changes, respectively. Conditions with $CR\geq0.50$ will be flagged as strongly coupled. Candidate pattern categories will be checked with and without high-drift or strongly coupled conditions; categories that change will be reported as drift-sensitive. The coupling-ratio threshold of 0.5 flags conditions where non-target drift exceeds half the intended change, representing a point at which the non-target metric shift is large enough to contaminate symptom attribution materially. Sensitivity will be checked using $\pm$0.10.

\subsection{Training-Dynamics Logging}

Following Shah et al.'s study on data bugs in deep learning models for software engineering~\cite{shah2025}, we will log training-time behaviour through error trajectories and model-internal statistics. Shah et al. use training logs, gradients, weights, and biases to study symptoms of data-quality and preprocessing issues in SE deep learning tasks~\cite{shah2025}. We adapt this idea to metric-based SDP by tracking four predefined metric families: (i) error dynamics, including training error, validation error, and the train--validation gap; (ii) gradient magnitude and propagation, including first-, middle-, and last-layer gradient RMS and the first-to-last gradient RMS ratio; (iii) gradient distribution shape, including skewness, kurtosis, and near-zero gradient proportion for selected layers; and (iv) parameter statistics, including selected weight and bias summaries.

Training and validation metrics will be logged at every epoch. Gradient summaries will be computed after backpropagation and before the optimizer update, then aggregated to epoch-level summaries. Run-level analyses will summarize epoch-level metrics using a fixed late-training window, operationalized as the final 20\% of epochs. This pragmatic choice reduces sensitivity to single-epoch noise while keeping the summary focused on late-training behaviour.

\subsection{Pattern Classification and Statistical Analysis}
We will treat all pattern categories as exploratory candidates, not causal diagnoses. Because multiple metrics are logged, isolated metric changes will not be interpreted as standalone evidence. For each dataset, intervention condition, seed, and logged metric, we will summarize epoch-level values using fixed window summaries described above. For each metric, we will compute the signed standardized change between the low-severity reference condition and higher-severity conditions. We will treat $|d|\geq0.5$ as a non-trivial standardized change, following the conventional medium-effect threshold for Cohen-style standardized effects~\cite{cohen_statistical_1988}. This floor is preferred over $|d|\geq0.2$ to avoid classifying substantively negligible changes as candidate patterns across the large number of metrics examined. Smaller effects will be reported descriptively but will not be used to assign exploratory pattern categories.

A candidate pattern will be considered direction-consistent if at least 70\% of eligible datasets show the same direction of change, while fewer than 20\% show the reverse direction. Datasets with negligible changes will be treated as neutral rather than opposite. The 70\% super-majority and 20\% reverse ceiling together ensure classification is not driven by a narrow dataset subset nor assigned when a non-trivial minority shows the opposite pattern.

Within a dataset, the pattern must also be seed-stable: at least four of the five training seeds must show the same sign of change. This near-unanimous criterion ensures the pattern is stable under within-condition stochastic variation in initialization and mini-batch ordering, rather than reflecting a single unlucky seed. Conditions with strong non-target drift will be flagged before classification. We define strong non-target drift as an absolute change of at least 0.10 in the non-target data-quality metric. Since C2 and N1 are bounded in $[0,1]$, this represents a 10-percentage-point change. The 0.10 drift threshold is used only as a within-metric operational flag for substantial change, not as a claim that C2 and N1 have equivalent substantive severity. We will therefore interpret drift using both the absolute within-metric change and the coupling ratio, and will report sensitivity analyses using 0.05 and 0.15 thresholds. Candidate pattern categories must remain qualitatively unchanged when high-drift conditions are excluded; otherwise, the pattern will be categorized as drift-sensitive.

We will use the following exploratory pattern categories. A pattern will be categorized as \emph{candidate imbalance-associated} if it satisfies the effect-size, direction-consistency, and seed-stability rules under imbalance-only intervention, but not under overlap-only intervention. A pattern will be categorized as \emph{candidate overlap-associated} using the analogous rule for overlap-only intervention. A pattern will be categorized as \emph{shared} if it satisfies the effect-size, direction-consistency, and seed-stability rules under both imbalance-only and overlap-only interventions, has the same direction of change, and has comparable magnitude. We operationalize comparable magnitude as a ratio of absolute standardized effects between 0.5 and 2.0, i.e., neither effect is more than twice the other. The 0.5–2.0 bounds are pre-specified to exclude asymmetric effects where one issue drives the pattern substantially more than the other; no theoretical derivation is claimed. For joint conditions, we compute an interaction residual by comparing the observed joint standardized change with the sum of the corresponding single-issue changes:
\[
\delta_m=d^{joint}_m(\alpha_I,\alpha_O)-[d^{imb}_m(\alpha_I)+d^{ov}_m(\alpha_O)].
\]
A pattern is categorized as \emph{interaction-dependent} when \(|\delta_m|\geq0.5\) and the direction-consistency and seed-stability rules are satisfied. A pattern will be categorized as \emph{masked/dominated} if a single-issue pattern falls below $|d|<0.2$ or changes direction when the other issue is introduced jointly. In this study, an informative outcome need not contain many stable pattern categories. A sparse or empty set of stable categories will be treated as evidence that the logged dynamics do not support robust issue-associated interpretation under this protocol. High fidelity failure, strong non-target drift, or unstable classifications will likewise be reported as evidence about protocol feasibility and limits.

\section{Threats to Validity}

\textbf{Construct validity.}
The interventions are controlled stressors rather than natural reproductions of SDP data-quality problems. Imbalance injection may reflect artefacts of random oversampling, undersampling, and changed class exposure; progressive imbalance reduction may reflect SMOTE artefacts because SMOTE creates synthetic minority instances rather than recovering natural data. We partly address this threats by reporting realized C2 values, fidelity checks, non-target drift, coupling ratios, class counts, and training-set sizes. Overlap injection may create label–feature tension, so overlap-associated patterns may partly reflect label–feature inconsistency rather than natural boundary ambiguity. We partly address this by: (i) treating the intervention as a controlled stressor, not a natural reproduction; (ii) using progressive reduction from natural overlap as bidirectional triangulation; and (iii) testing sensitivity to $\gamma=0.60$.

\textbf{Internal and conclusion validity.}
Training dynamics may be influenced by mini-batch composition, local density changes, training-set size, stochastic optimization, and the fixed MLP architecture, not only by the targeted data-quality issue. N1 may also be unstable in small or high-dimensional datasets because it relies on nearest-neighbour structure. 

\textbf{External validity and reproducibility.}
The findings will be limited to metric-based SDP datasets and one neural learner; they should therefore be interpreted as candidate patterns for this setting rather than universal symptoms of imbalance or overlap. Results may not transfer to just-in-time defect prediction, process metrics, or code-token models. Reproducibility may also be affected by software versions, hardware, CUDA operations, and framework nondeterminism, so we will report package versions, device information, seeds, configurations, intervention logs, and realized data-quality measurements.

\section{Conclusion}

This study proposes an interaction-aware empirical protocol for examining how coupled data-quality issues may shape neural training behaviour in software defect prediction. The protocol systematically manipulates class imbalance and class overlap, both separately and jointly, while monitoring whether an intervention targeting one issue unintentionally changes the other. In this way, the study moves beyond endpoint performance evaluation by examining training-time behaviour, including gradients, weights, biases, and training/validation error trajectories.

Rather than treating observed training-dynamics patterns as issue-specific by default, the protocol categorizes them as candidate imbalance-associated, overlap-associated, shared, interaction-dependent or masked/dominated patterns. The resulting evidence of the final study will provide a transparent and reproducible basis for understanding how coupled data-quality issues affect neural software defect prediction models and may inform future diagnostic tools for software maintenance and evolution.

\section*{Declaration of AI Assistance}
The authors used ChatGPT to support language editing. The authors reviewed, revised, and take responsibility for all content.

\bibliographystyle{IEEEtran}
\bibliography{IEEEabrv,mybibfile}

\end{document}